\documentclass[conference]{IEEEtran}
\IEEEoverridecommandlockouts
\usepackage{cite}
\usepackage{amsmath,amssymb,amsfonts}
\usepackage{graphicx}
\usepackage{textcomp}
\usepackage{xcolor}
\usepackage{booktabs}
\usepackage{url}
\usepackage{hyperref}
\usepackage{tikz}
\usepackage{float}
\usetikzlibrary{arrows.meta,positioning,calc,fit,backgrounds,decorations.pathreplacing}

\begin{document}

\title{Weather-Robust Scene Semantics with Vision-Aligned 4D Radar}

\author{%
\IEEEauthorblockN{Kali Hamilton and Christoffer Heckman}
\thanks{Authors are with the Autonomous Robotics and Perception Group at the University of Colorado Boulder, Boulder, CO 80309, USA.}%
}
\maketitle

\begin{abstract}
Cameras and LiDAR degrade in rain, fog, and snow, while millimeter-wave
radar remains largely unaffected. We align a radar encoder to frozen
SigLIP vision embeddings and decode structured scene captions through a
frozen vision-language model (VLM) with ${\sim}$7M trainable parameters. On
K-RADAR~\cite{kradar} with held-out fog, light snow, and heavy snow
sequences, all radar configurations outperform a camera baseline that
collapses to $>$90\% hallucination. We identify a token-norm mismatch as
the dominant failure mode when bridging radar to a frozen VLM and show
that projector-output LayerNorm resolves it. Analysis of encoder
complexity, caption format, and pooling strategy reveals tradeoffs that
inform future radar--VLM pipeline design.
\end{abstract}


% ============================================================================
\section{Introduction}
% ============================================================================

Mobile robots operating outdoors routinely face conditions that defeat cameras
and LiDAR: rain, fog, dust, snow, and airborne particulates.
Millimeter-wave radar at 77~GHz penetrates these conditions, and modern 4D
imaging radar produces a dense $(\mathrm{range},\mathrm{azimuth},
\mathrm{elevation},\mathrm{Doppler})$ tensor rich enough to motivate learned
scene understanding rather than hand-crafted detection alone.

Vision-language models (VLMs) have demonstrated strong open-vocabulary scene
understanding by aligning a vision encoder to a large language model~\cite{llava}.
Robotics work such as RT-2~\cite{rt2} and $\pi_0$~\cite{pi0} shows that the
same pattern (a generic CNN adapted for the sensor feeding a frozen foundation
model) transfers across modalities. We ask: \emph{can a radar encoder be
aligned to a vision embedding space, and can a frozen VLM decode the resulting
tokens into structured scene descriptions?} Rather than building a
fixed-class detection head, we use structured caption generation as a proxy
to evaluate the quality of the radar--vision alignment under adverse weather.

Our contributions are:
\begin{itemize}
  \item A \textbf{caption generation and evaluation pipeline} for
    K-RADAR~\cite{kradar}: programmatic ground-truth captions from 3D
    bounding boxes, structured parsing, and caption-as-detection metrics
    enabling weather-stratified evaluation of radar--VLM systems.
  \item \textbf{Empirical analysis of vision--radar alignment} at low
    data scale ($\sim$8k frames), including an encoder--VLM utilization
    gap, caption format tradeoffs (prose vs.\ JSON), and pooling
    strategy effects on spatial prediction.
\end{itemize}

% ============================================================================
\section{Related Work}
% ============================================================================

\textbf{4D radar perception.}
K-RADAR~\cite{kradar} is the first large-scale 4D imaging-radar dataset with
3D bounding-box annotations across weather conditions. RADDet~\cite{raddet}
treats the Doppler dimension as CNN input channels for detection, motivating
our 66-channel input variant. RTNH~\cite{rtnh} and L4DR~\cite{l4dr} demonstrate
4D radar's weather robustness for detection; ColoRadar~\cite{coloradar} extends
millimeter-wave radar to robot platforms beyond on-road settings. We use structured captioning as a proxy to evaluate radar encoder
alignment quality under adverse weather.

\textbf{Radar-language models.}
RSLM~\cite{rslm} aligns a 2D radar-spectrum encoder to a frozen VLM via
CLIP-style contrastive loss. RLM~\cite{rlm} trains a generative captioner on
800k CARLA-simulated radar-caption pairs. Talk2Radar~\cite{talk2radar} publishes
a real-4D-radar referring-expression dataset with a cross-modal fusion head
rather than a frozen LLM. Our work targets the unaddressed combination:
generative captioning from \emph{real} 4D radar tensors against a frozen VLM.
Simulated radar is, in practice, a ray-cast of LiDAR geometry with a
noise model, erasing the multipath and precipitation phenomena that define
radar's weather advantage.

\textbf{Position encoding for metric sensors.}
PETR~\cite{petr} injects metric coordinates additively into intermediate
feature maps, avoiding the dilution that CoordConv-style input channels
suffer after strided convolutions. We adopt PETR after \texttt{layer2}.

% ============================================================================
\section{Method}
% ============================================================================

\subsection{Input Representation}
\begin{figure}[t]
\centering
\resizebox{\columnwidth}{!}{% Tesseract -> elevation projection -> 5ch vs 66ch input (compact)
\begin{tikzpicture}[
    box/.style={draw, rounded corners=2pt, minimum height=0.8cm,
                text width=2.0cm, align=center, font=\small},
    arr/.style={-{Stealth[length=4pt]}, thick},
    label/.style={font=\scriptsize, text=black!70},
  ]

  % 4D Tesseract
  \node[box, fill=orange!15, text width=2.4cm] (tess)
    {4D Tesseract\\[1pt]
     \scriptsize$(64{\times}256{\times}37{\times}107)$\\[-1pt]
     \scriptsize Dop$\times$R$\times$El$\times$Az};

  % Elevation projection
  \node[box, fill=yellow!15, right=0.8cm of tess, text width=2.0cm] (elproj)
    {El.\ Max-Proj\\[1pt]
     \scriptsize$(64{\times}256{\times}107)$\\[-1pt]
     \scriptsize Dop$\times$R$\times$Az};

  \draw[arr] (tess) -- (elproj)
    node[midway, above=1pt, label] {$\max$ El};

  % R^4 compensation
  \node[box, fill=yellow!15, right=0.6cm of elproj, text width=1.8cm] (r4)
    {$R^4$ Comp.\\[1pt]
     \scriptsize range power\\[-1pt]
     \scriptsize correction};

  \draw[arr] (elproj) -- (r4);

  % Branch point
  \coordinate (branch) at ($(r4.east)+(0.5,0)$);
  \draw[thick] (r4.east) -- (branch);

  % --- 5ch branch (top) ---
  \node[box, fill=green!12, text width=2.2cm,
        above right=0.6cm and 0.3cm of branch] (agg)
    {Doppler Agg.\\[1pt]
     \scriptsize $64{\to}3$: power,\\[-1pt]
     \scriptsize mean, peak};

  \draw[arr] (branch) |- (agg);

  \node[box, fill=green!25, text width=2.6cm, right=0.5cm of agg] (fivech)
    {\textbf{5ch} $[5,256,107]$\\[1pt]
     \scriptsize 3 Doppler + range + az};

  \draw[arr] (agg) -- (fivech);

  % --- 66ch branch (bottom) ---
  \node[box, fill=blue!12, text width=2.2cm,
        below right=0.6cm and 0.3cm of branch] (keep)
    {All Bins\\[1pt]
     \scriptsize full velocity\\[-1pt]
     \scriptsize distribution};

  \draw[arr] (branch) |- (keep);

  \node[box, fill=blue!25, text width=2.6cm, right=0.5cm of keep] (sixsixch)
    {\textbf{66ch} $[66,256,107]$\\[1pt]
     \scriptsize 64 Doppler + range + az};

  \draw[arr] (keep) -- (sixsixch);

\end{tikzpicture}}
\caption{Input representation: 4D tesseract $\to$ elevation max-projection
$\to$ $R^4$ compensation $\to$ 5-channel or 66-channel variants.}
\label{fig:input_repr}
\end{figure}
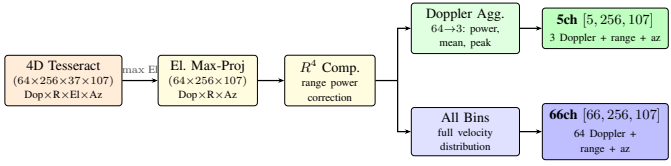

Each K-RADAR frame is a 4D tesseract of shape $(64, 256, 37, 107)$ in
$(\text{Doppler}, \text{range}, \text{elevation}, \text{azimuth})$. Because
the radar's vertical resolution is coarse (37 bins over $\pm15^\circ$) and
traffic objects occupy a narrow height band, we collapse elevation by taking
the maximum power across all 37 bins at each $(\text{Doppler},
\text{range}, \text{azimuth})$ cell, yielding a 3D tensor of shape
$(64, 256, 107)$. This max-projection is standard for 4D radar:
RADDet~\cite{raddet} and RTNH~\cite{rtnh} apply the same reduction to
obtain range-azimuth-Doppler volumes, and K-RADAR's own
baselines~\cite{kradar} operate on elevation-collapsed slices.
We then apply $R^4$ range compensation to correct free-space path loss
and produce two input variants (Fig.~\ref{fig:input_repr}):

\begin{itemize}
\item \textbf{5-channel} $[5, 256, 107]$: total $R^4$-compensated power,
  mean Doppler velocity, peak Doppler velocity, physical range~(m), and
  azimuth~(deg). This compact, physics-informed summary captures the
  dominant velocity and spatial structure per cell.
\item \textbf{66-channel} $[66, 256, 107]$: all 64 Doppler bins with $R^4$
  compensation, plus range and azimuth coordinate channels. Preserving
  the full velocity distribution per cell retains multi-target and
  micro-Doppler information that the 5-channel aggregation discards.
\end{itemize}

\noindent Both variants append two metric coordinate channels (range in
metres, azimuth in degrees). Fig.~\ref{fig:input_repr} summarizes
the pipeline.

\subsection{Architecture}
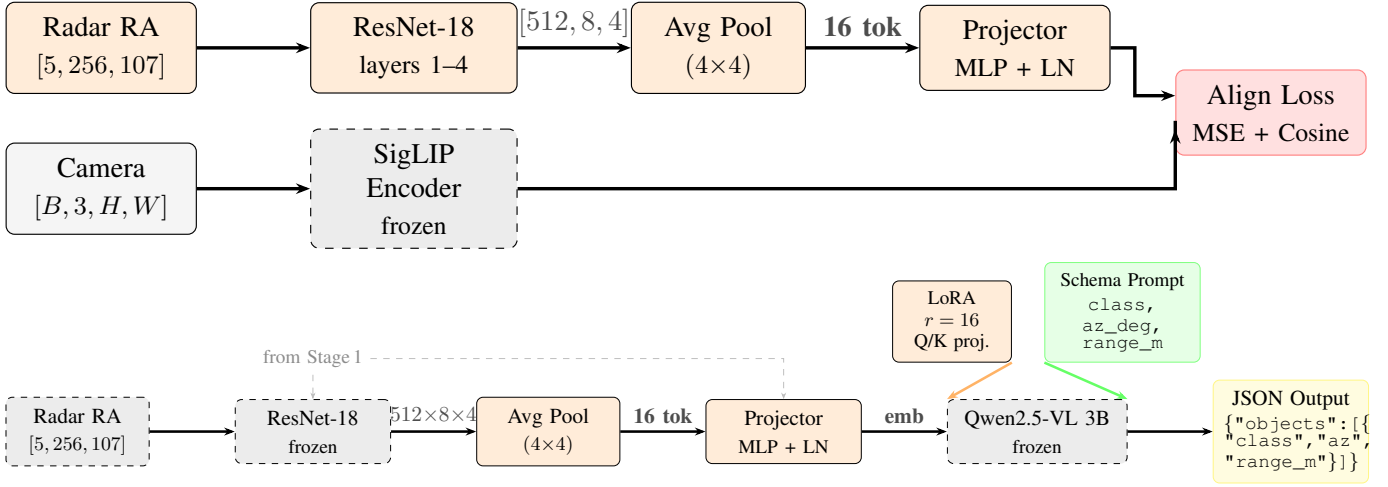
\begin{figure*}[t]
\centering
\resizebox{\textwidth}{!}{% Stage 1: Vision-radar alignment (compact, no trainable/frozen on camera)
\begin{tikzpicture}[
    box/.style={draw, rounded corners=3pt, minimum height=1.0cm,
                align=center, font=\normalsize, inner sep=5pt},
    frozen/.style={box, fill=gray!15, dashed},
    train/.style={box, fill=orange!15},
    loss/.style={box, fill=red!12, draw=red!50},
    arr/.style={-{Stealth[length=5pt]}, very thick},
    dim/.style={font=\normalsize\bfseries, text=black!70},
    node distance=1.4cm,
  ]

  % === Radar path (top) ===
  \node[train, text width=2.0cm] (rinput)
    {Radar RA\\[1pt]\small$[5,256,107]$};

  \node[train, text width=2.2cm, right=1.4cm of rinput] (enc)
    {ResNet-18\\[1pt]\small layers 1--4};

  \node[train, text width=1.8cm, right=1.4cm of enc] (pool)
    {Avg Pool\\[1pt]\small$(4{\times}4)$};

  \node[train, text width=2.0cm, right=1.4cm of pool] (proj)
    {Projector\\[1pt]\small MLP + LN};

  \draw[arr] (rinput) -- (enc);
  \draw[arr] (enc) -- node[above, dim] {$[512,8,4]$} (pool);
  \draw[arr] (pool) -- node[above, dim] {16 tok} (proj);

  % === Camera path (bottom) ===
  \node[box, fill=gray!8, text width=2.0cm, below=0.6cm of rinput] (cam)
    {Camera\\[1pt]\small$[B,3,H,W]$};

  \node[frozen, text width=2.2cm, right=1.4cm of cam] (siglip)
    {SigLIP\\Encoder\\[1pt]\small frozen};

  \draw[arr] (cam) -- (siglip);

  % === Loss node ===
  \node[loss, text width=2.0cm, right=0.8cm of proj, yshift=-0.8cm] (loss)
    {Align Loss\\[1pt]\small MSE + Cosine};

  \draw[arr] (proj.east) -- ++(0.3,0) |- (loss.170);
  \coordinate (sigturn) at (siglip.east -| loss.west);
  \draw[arr] (siglip.east) -- (sigturn) |- (loss.190);

\end{tikzpicture}}\\[0.5em]
\resizebox{\textwidth}{!}{% Stage 2: Radar-VLM captioning pipeline (compact, no legend)
\begin{tikzpicture}[
    box/.style={draw, rounded corners=3pt, minimum height=1.0cm,
                align=center, font=\normalsize, inner sep=5pt},
    frozen/.style={box, fill=gray!15, dashed},
    train/.style={box, fill=orange!15},
    prompt/.style={box, fill=green!10, draw=green!50},
    output/.style={box, fill=yellow!15, draw=yellow!60},
    arr/.style={-{Stealth[length=5pt]}, very thick},
    dim/.style={font=\normalsize\bfseries, text=black!70},
    note/.style={font=\small, text=black!45, align=center},
    node distance=1.4cm,
  ]

  % ===== Main pipeline =====
  \node[frozen, text width=2.0cm] (input)
    {Radar RA\\[1pt]\small$[5,256,107]$};

  \node[frozen, text width=2.2cm, right=1.4cm of input] (enc)
    {ResNet-18\\[1pt]\small frozen};

  \node[train, text width=2.0cm, right=1.4cm of enc] (xpool)
    {Avg Pool\\[1pt]\small$(4{\times}4)$};

  \node[train, text width=2.2cm, right=1.4cm of xpool] (proj)
    {Projector\\[1pt]\small MLP + LN};

  \node[frozen, text width=2.6cm, right=1.4cm of proj] (vlm)
    {Qwen2.5-VL 3B\\[1pt]\small frozen};

  \node[output, text width=2.2cm, right=1.4cm of vlm] (out)
    {JSON Output\\[1pt]
     \small\texttt{\{"objects":[\{}\\[-2pt]
     \small\texttt{"class","az",}\\[-2pt]
     \small\texttt{"range\_m"\}]\}}};

  % Main flow arrows
  \draw[arr] (input) -- (enc);
  \draw[arr] (enc) -- node[above, dim] {$512{\times}8{\times}4$} (xpool);
  \draw[arr] (xpool) -- node[above, dim] {16 tok} (proj);
  \draw[arr] (proj) -- node[above, dim] {emb} (vlm);
  \draw[arr] (vlm) -- (out);

  % ===== LoRA badge (above-left of VLM) =====
  \node[train, minimum height=0.4cm, text width=1.6cm,
        font=\small, above=0.6cm of vlm, xshift=-1.4cm] (lora)
    {LoRA $r\!=\!16$\\Q/K proj.};
  \draw[arr, draw=orange!60] (lora.south east) -- (vlm.north west);

  % ===== Schema prompt (above-right of VLM) =====
  \node[prompt, minimum height=0.4cm, text width=2.2cm,
        font=\small, above=0.6cm of vlm, xshift=1.4cm] (prompt)
    {Schema Prompt\\[1pt]\small\texttt{class, az\_deg,}\\[-2pt]\small\texttt{range\_m}};
  \draw[arr, draw=green!60] (prompt.south west) -- (vlm.north east);

  % ===== Weight transfer from Stage 1 =====
  \node[note, above=0.4cm of enc] (wt) {from Stage\,1};
  \draw[-{Stealth[length=3pt]}, black!25, dashed]
    (wt.south) -- (enc.north);
  \draw[-{Stealth[length=3pt]}, black!25, dashed]
    (wt.east) -| (proj.north);

\end{tikzpicture}}
\caption{\textbf{Two-stage training pipeline.}
\emph{Top:} Stage~1 aligns the radar encoder to frozen SigLIP~\cite{siglip}
vision embeddings (MSE + cosine loss).
\emph{Bottom:} Stage~2 freezes the encoder, transfers projector weights, and
fine-tunes LoRA~\cite{lora} adapters on Qwen2.5-VL-3B~\cite{qwen25vl} to
generate structured captions from radar tokens.}
\label{fig:architecture}
\end{figure*}

The pipeline consists of two stages (Fig.~\ref{fig:architecture}):
vision--radar alignment, followed by caption fine-tuning.

\textbf{Radar encoder.} A standard ResNet-18 with its first convolution
reshaped to accept $C \in \{5, 66\}$ input channels. After
\texttt{layer2} (stride 8, 128 channels, $32{\times}14$ feature map), we
inject a PETR-style additive position embedding: normalized
$(\text{range}, \text{azimuth})$ coordinates are mapped through a 2-layer
MLP to 128 channels and added element-wise. After \texttt{layer4}
($512{\times}8{\times}4$), average pooling with a $4{\times}4$ kernel
produces 16 tokens of 512 dimensions. We also ablate a cross-attention
pool variant with 32 learned queries (Table~\ref{tab:detection},
rightmost column).

\textbf{Projector with output LayerNorm.} A 2-layer MLP
$(512 \to 2048 \to 2048)$ with GELU maps the 32 radar tokens into the VLM
embedding dimension. We use an MLP rather than a Q-Former: at our data scale
(7k frames, no BLIP-2 pretraining), MLP projectors are the standard
sensor-to-LLM choice~\cite{pointllm,lamm}.

We apply LayerNorm to the projector \emph{output}. The failure
mode that motivated this is subtle and easily missed: training loss
decreased normally over many epochs, but a swap test (substituting
zeros or random noise for the radar tensor) produced identical
captions. The VLM had learned to generate from its language prior
while routing attention around the radar tokens. Diagnosing this required
explicitly comparing output norms: projector outputs had mean L2 norm
$\sim$18--207, while Qwen's native embedding matrix converges near unit
norm under its joint RMSNorm training. With a $20{\times}$ norm advantage,
radar tokens saturate the softmax and collect attention weight in the forward
pass; but the resulting attention distribution carries near-zero variance
over radar positions, starving those inputs of gradient and reinforcing the
pathology. Output LayerNorm resolves both the forward-pass saturation and
the gradient starvation in one operation. Q-Former architectures handle this implicitly through their
internal normalization; using a simple MLP projector exposes the
mismatch and requires an explicit fix.

\textbf{Frozen VLM with LoRA.} Qwen2.5-VL-3B-Instruct~\cite{qwen25vl} is
loaded in fp16 and frozen. Radar
token embeddings are prepended to the text prompt, and the model generates
the caption autoregressively. Rank-16 LoRA~\cite{lora} adapters on the
$Q/K$ attention projections across all 36 transformer layers (3.7M
trainable parameters) allow the frozen backbone to mildly adapt its
attention distribution to radar tokens. Loss is standard cross-entropy on
caption tokens only.

\subsection{Caption Generation}\label{sec:captions}

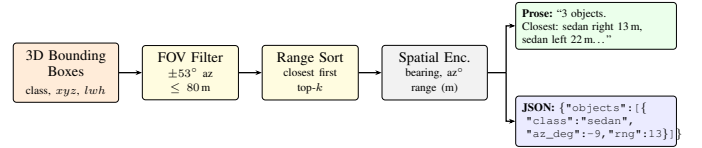
\begin{figure}[H]
\centering
\resizebox{\columnwidth}{!}{% Caption generation pipeline: 3D bbox -> prose / JSON captions (compact)
\begin{tikzpicture}[
    box/.style={draw, rounded corners=2pt, minimum height=0.8cm,
                text width=1.8cm, align=center, font=\small, inner sep=4pt},
    outbox/.style={draw, rounded corners=2pt, minimum height=0.7cm,
                   text width=3.2cm, align=left, font=\scriptsize, inner sep=4pt},
    arr/.style={-{Stealth[length=4pt]}, thick},
    label/.style={font=\scriptsize, text=black!60},
  ]

  % --- Pipeline boxes ---
  \node[box, fill=orange!15, text width=2.0cm] (bbox)
    {3D Bounding\\Boxes\\[1pt]
     \scriptsize class, $xyz$, $lwh$};

  \node[box, fill=yellow!15, right=0.5cm of bbox] (fov)
    {FOV Filter\\[1pt]
     \scriptsize $\pm53^\circ$ az\\[-1pt]
     \scriptsize $\leq80$\,m};

  \draw[arr] (bbox) -- (fov);

  \node[box, fill=yellow!15, right=0.5cm of fov] (sort)
    {Range Sort\\[1pt]
     \scriptsize closest first\\[-1pt]
     \scriptsize top-$k$};

  \draw[arr] (fov) -- (sort);

  \node[box, fill=gray!10, right=0.5cm of sort, text width=2.0cm] (spatial)
    {Spatial Enc.\\[1pt]
     \scriptsize bearing, az$^\circ$\\[-1pt]
     \scriptsize range (m)};

  \draw[arr] (sort) -- (spatial);

  % --- Branch ---
  \coordinate (branch) at ($(spatial.east)+(0.4,0)$);
  \draw[thick] (spatial.east) -- (branch);

  % --- Prose output ---
  \node[outbox, fill=green!8, above right=0.5cm and 0.2cm of branch] (prose)
    {\textbf{Prose:} ``3 objects.\\
     Closest: sedan right 13\,m,\\
     sedan left 22\,m\ldots''};

  \draw[arr] (branch) |- (prose);

  % --- JSON output ---
  \node[outbox, fill=blue!8, below right=0.5cm and 0.2cm of branch] (json)
    {\textbf{JSON:} \texttt{\{"objects":[\{}\\
     \texttt{~"class":"sedan",}\\
     \texttt{~"az\_deg":-9,"rng":13\}]\}}};

  \draw[arr] (branch) |- (json);

\end{tikzpicture}}
\caption{Caption generation pipeline. LiDAR-derived 3D bounding-box
annotations are filtered to the radar FOV, sorted by range, and
formatted as prose or JSON ground truth.}
\label{fig:caption_pipeline}
\end{figure}

Ground-truth captions are generated programmatically from K-RADAR's 3D
bounding-box annotations, which are derived from LiDAR
(Fig.~\ref{fig:caption_pipeline}).
Each frame's bounding boxes are filtered to the radar FOV
($\pm53^\circ$ azimuth, $\leq80$\,m range), sorted by range, and the
closest objects receive per-object descriptions with class, range~(m),
and azimuth (as a bearing sector or numeric degrees).

\textbf{Prose} captions use natural-language templates:
\emph{``There are 3 objects. Closest: a sedan slightly to the right
at 13\,m, a sedan to the left at 22\,m\ldots''}

\textbf{JSON} captions encode each object as a structured dictionary:
\texttt{\{"objects":[\{"class":"sedan",\allowbreak "azimuth\_deg":-9,\allowbreak "range\_m":13\}]\}}.
Numeric fields provide direct digit-token supervision for range and
azimuth. This format trades spatial precision for recall
(Section~\ref{sec:format}).

\subsection{Training}
\textbf{Stage 1: Vision alignment.} The radar encoder and projector are
jointly trained to regress frozen SigLIP~\cite{siglip} features on paired
radar-camera frames using MSE plus cosine distance loss.

\textbf{Stage 2: Caption fine-tuning.} The encoder is frozen. The projector
is initialized from Stage 1 and fine-tuned jointly with rank-16 LoRA on
Qwen's $Q/K$ projections using cross-entropy on caption tokens.

\subsection{Dataset and Split}
We use a sequence-level split of K-RADAR~\cite{kradar} to avoid temporal
frame leakage: 12 training sequences ($\sim$7.5k frames), 3 validation,
and 4 test. Fog and light snow sequences are fully held out from training,
giving zero-shot weather evaluation. The full per-sequence breakdown is in
Table~\ref{tab:dataset} (Appendix).

% ============================================================================
\section{Results}
% ============================================================================

\subsection{Caption-as-Detection Metrics}

We parse generated captions into structured predictions (class, range in
meters, angular position) and score against caption-parsed ground truth.
\textbf{Class P/R/F1} uses multiset matching on class per frame. \textbf{Range MAE} is computed over matched pairs; because range is
a text token, this is a comparative metric across configurations, not
physical sensor precision. Prose models output one of 7 \textbf{bearing
sectors}; JSON models output integer azimuth in degrees, evaluated as
\textbf{azimuth MAE}. \textbf{Hallucination rate} is the fraction of
predicted classes absent from any GT object in the same frame.

Table~\ref{tab:detection} shows results on held-out test sequences
(fog, light snow, heavy snow). No single configuration wins every metric.
The camera baseline uses the same VLM and LoRA setup but with an unfrozen
ResNet-18 encoder on camera input (no Stage-1 alignment), so it measures
VLM capability on a familiar modality rather than a controlled
sensor-only comparison.

\begin{table}[H]
\caption{Detection metrics on held-out weather sequences. Bold = best per row.}
\label{tab:detection}
\centering
\small
\setlength{\tabcolsep}{3pt}
\begin{tabular}{@{}lccccc@{}}
\toprule
 & \textit{Baseline} & \multicolumn{4}{c}{\textit{Ours}} \\
\cmidrule(lr){2-2}\cmidrule(lr){3-6}
 & \textbf{Camera} & \textbf{5ch} & \textbf{66ch} & \textbf{5ch JSON} & \textbf{5ch JSON} \\
 & \textit{unfrozen} & \textit{+align} & \textit{+align} & \textit{avg pool} & \textit{x-attn} \\
\midrule
Class F1          & 0.336 & 0.527 & 0.473 & \textbf{0.540} & 0.491 \\
Precision         & 0.457 & \textbf{0.704} & 0.624 & 0.672 & 0.664 \\
Recall            & 0.266 & 0.421 & 0.381 & \textbf{0.452} & 0.390 \\
Range MAE (m)     & 17.0  & 13.9  & \textbf{10.9}  & 16.6  & 13.2 \\
Bearing acc.      & ---   & \textbf{0.333} & 0.304 & ---   & ---   \\
Azimuth MAE ($^\circ$) & 9.3 & --- & --- & \textbf{7.0} & 7.9 \\
Hallucination     & 0.543 & \textbf{0.296} & 0.376 & 0.329 & 0.336 \\
\bottomrule
\end{tabular}
\end{table}

\subsection{Weather Robustness}\label{sec:weather}

\begin{figure}[H]
\centering
\includegraphics[width=\columnwidth]{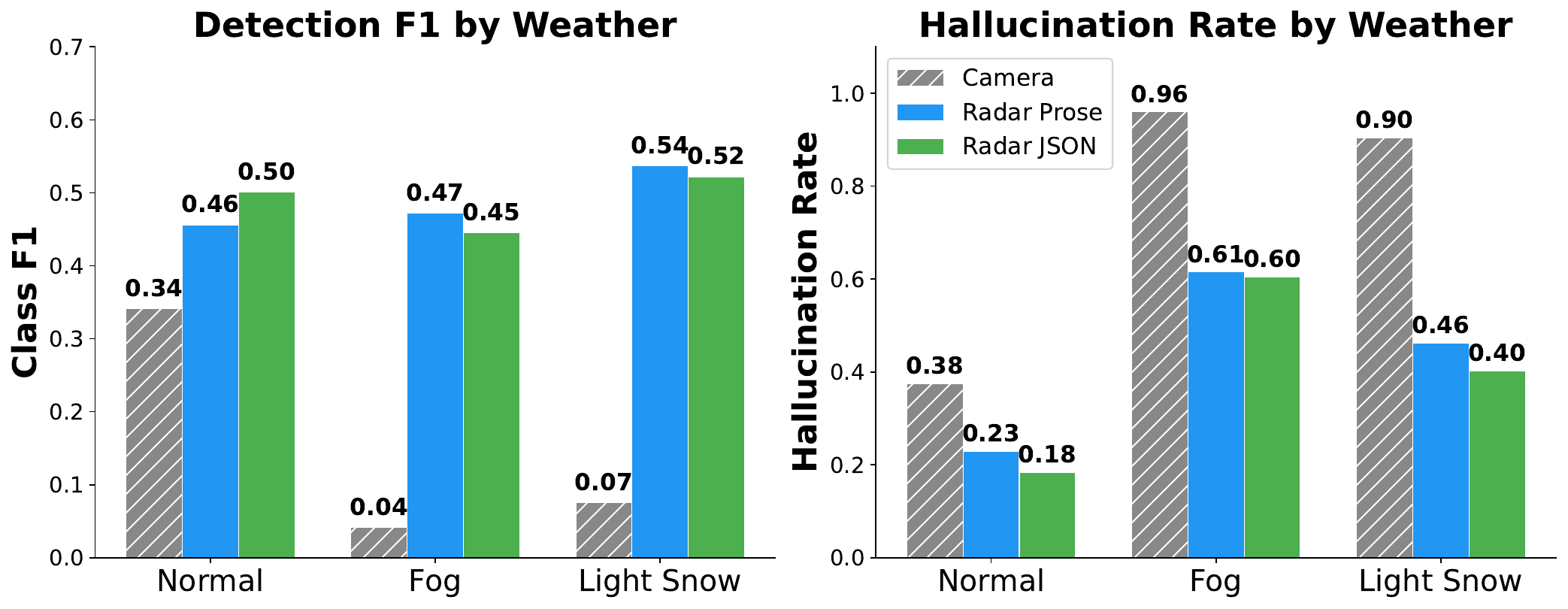}
\caption{\textbf{Per-weather detection F1 and hallucination rate.}
Camera collapses in fog and snow ($>$90\% hallucination);
both radar configurations maintain consistent detection.}
\label{fig:weather_bar}
\end{figure}

Fig.~\ref{fig:weather_bar} stratifies detection F1 and hallucination rate
by weather condition. The fog and light snow sequences are
entirely held out from Stage-1 and Stage-2 training.
Camera performance collapses to near-zero F1 in fog (0.04) and light snow
(0.07) with hallucination rates exceeding 90\%, while both radar models
maintain F1 above 0.44 across all conditions.
Fig.~\ref{fig:qualitative} (Appendix) shows qualitative examples
across normal, fog, and heavy snow conditions.

% ============================================================================
\section{Discussion}\label{sec:discussion}
% ============================================================================

\subsection{Why VL, Not Text-Only}
In earlier ablations we swapped Qwen2.5-VL for a text-only Qwen2.5 of
similar scale while holding the encoder, projector, and LoRA fixed. The
text-only variant underperformed the VL variant by 8\% on validation loss, suggesting
that vision-conditioned attention patterns from VLM pretraining transfer
usefully to radar tokens. The VLM's cross-modal attention appears to generalize beyond the
specific pixel statistics it was trained on.

\subsection{What the Configurations Tell Us}

\textbf{Alignment matters for spatial grounding.} In early ablations,
training without Stage-1 SigLIP alignment produced encoders that learned
class statistics but failed to ground spatial predictions, resulting in
high hallucination and poor bearing accuracy. The latent space
misalignment between a randomly-initialized encoder and the frozen VLM
was too large to bridge with caption loss alone at this data scale.
Stage-1 alignment closes this gap, enabling all aligned radar models to
achieve 7.0--7.9$^\circ$ azimuth MAE with substantially lower
hallucination (0.30--0.38 vs.\ 0.54 for the unaligned camera baseline).

\textbf{5ch wins captions; 66ch wins encoder features.} 5ch aligned leads on
F1, precision, bearing, and hallucination; 66ch aligned leads on range MAE
(10.9~m vs.\ 13.9~m). To disentangle encoder quality from VLM utilization, we
freeze each trained encoder and fit linear probes on the pooled 512-dim
features. On object counting, 66ch achieves RMSE 2.03 vs.\ 2.20 for 5ch; on
bus/truck presence, 66ch reaches 84.2\% vs.\ 81.9\%. The full Doppler spectrum produces better encoder representations,
yet the frozen VLM generates worse captions, suggesting that alignment
difficulty scales with encoder complexity, particularly under limited
training data ($\sim$8k frames). The simpler 5-channel input aligns more
efficiently at this data scale; with more data or a richer projector,
66ch should surpass 5ch.

\begin{figure}[t]
\centering
\includegraphics[width=\columnwidth]{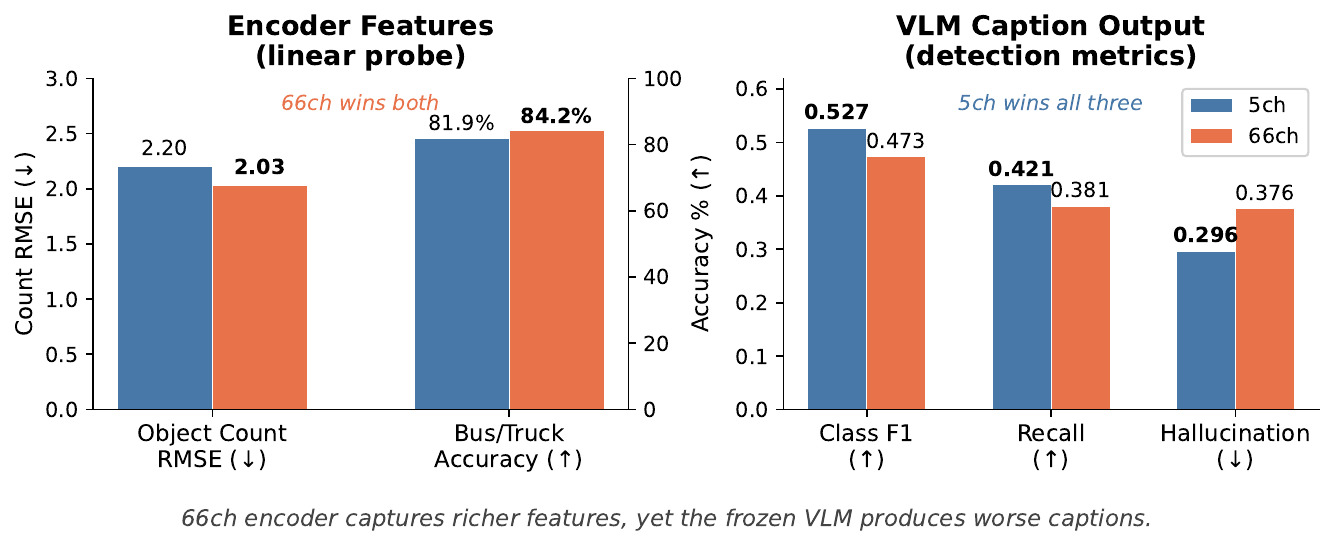}
\caption{Encoder--VLM utilization gap. Linear probes on frozen encoder
features (left) show 66ch captures richer representations; yet the
frozen VLM produces worse captions (right), suggesting alignment
difficulty scales with encoder complexity under limited data.}
\label{fig:utilization_gap}
\end{figure}

\textbf{Hallucination tracks alignment and complexity.} The aligned 5ch
model achieves the lowest hallucination (0.296), while the more complex
66ch aligned model returns to 0.376 despite richer encoder features.
When the encoder is over-parameterized for the data scale, the VLM's
language prior fills in plausible objects the radar signal cannot support.
Alignment on a compact representation is the strongest lever against
hallucination at this data scale.

\subsection{Caption Format as a Design Axis}\label{sec:format}

The JSON configurations in Table~\ref{tab:detection} reveal that caption
format is itself a meaningful design choice. JSON with average pooling
achieves the highest F1 (0.540) and recall (0.452), outperforming all prose
models, because numeric values appear as literal digit tokens: the model
learns to associate radar features with digit outputs directly rather than
bridging through natural-language phrases. Average pooling also produces the
best azimuth MAE (7.0$^\circ$), showing that spatial information survives
pooling when the output format provides direct numeric supervision.

Replacing average pooling with cross-attention pooling (32 learned queries)
improves range MAE (16.6$\to$13.2~m) but degrades F1 (0.540$\to$0.491)
and azimuth MAE (7.0$\to$7.9$^\circ$). The range--azimuth tradeoff across
pooling strategies suggests a competition for the model's numeric token
capacity: autoregressive generation processes spatial fields sequentially,
and the model allocates learning capacity unevenly across fields depending
on generation order and pooling structure. This is an artifact of using
language priors for spatial prediction rather than a sensor limitation.

JSON hallucination (0.329--0.336) remains higher than aligned prose (0.296).
Prose detects bicycle and motorcycle where JSON does not, suggesting
natural-language templates better support minority classes at low data
scale.

\subsection{Limitations}

Our best test-set F1 (0.540 for JSON, 0.527 for prose) is not
competitive with dedicated 4D-radar detection pipelines on localization,
nor with camera VLMs on clear-weather richness. The contribution is
demonstrating that vision-aligned radar encoders produce semantic
representations usable by a frozen VLM, and that the pipeline persists
through adverse weather where camera-based approaches fail entirely.

\textbf{Dataset scale.} Twelve training sequences ($\sim$7.5k frames) is small
by VLM standards, and our weather stratification holds out only one sequence
per adverse condition. Per-condition F1 values are point estimates over a
single route. Fig.~\ref{fig:weather_bar} should be read as: captioning
\emph{persists} through adverse weather on at least these sequences, not
that the pipeline has learned a weather-invariant representation.

\textbf{Evaluation ceiling from caption format.} GT captions detail at most
$\sim$4 objects per frame regardless of scene density. On 16-object scenes
the recall ceiling from the GT format is lower than the recall ceiling from
the model, so absolute recall numbers are a lower bound on what a
structured-output head would achieve.

\textbf{Autoregressive spatial prediction.} Range and azimuth are
predicted as text tokens, not regressed as scalars. Range MAE
(10.9--16.6~m) far exceeds the sensor's sub-meter resolution, and
adjacent meter values may share or split BPE pieces non-monotonically.
A regression head would bypass this tokenization ceiling entirely,
suggesting that language generation may not be the right decoder for
metric spatial fields.

% ============================================================================
\section{Conclusion}
% ============================================================================

Radar has a dearth of labeled data compared to camera and LiDAR, making
large-scale supervised training impractical. By aligning a radar encoder to
frozen vision embeddings and decoding through a frozen VLM, we get
structured scene understanding from $\sim$8k frames and $\sim$7M trainable
parameters, with no radar-specific language data collection. On held-out
fog, light snow, and heavy snow sequences, radar-based captioning persists
where camera input degrades or collapses entirely, and a single architectural fix (projector-output
LayerNorm) determines whether the VLM attends to radar at all. An
encoder--VLM utilization gap reveals that richer encoder features do not
automatically yield better captions under limited data, highlighting
alignment efficiency as a key bottleneck at this scale.

Our results also show that autoregressive token prediction introduces
arbitrary tradeoffs between spatial fields depending on generation order,
and absolute spatial errors remain far above sensor resolution. A
regression head or detection decoder over the same aligned embeddings
would bypass these tokenization artifacts and provide a cleaner evaluation
of encoder quality. LiDAR-as-teacher alignment could provide richer 3D
supervision than camera, and cross-dataset transfer
(e.g.\ ColoRadar~\cite{coloradar}) would test whether the learned
alignment generalizes beyond a single sensor platform.

% ============================================================================
\section*{Acknowledgments}
We thank the K-RADAR team for releasing raw 4D radar tensors across diverse
weather conditions.

\textbf{AI use disclosure.} Claude (Anthropic) was used for copy-editing
and prose polishing; all technical content, results, and analysis are the
authors' own. The pipeline architecture diagram was generated with Gemini
(Google) from an author-written specification and then hand-corrected.

% ============================================================================
\bibliographystyle{IEEEtran}
\bibliography{references}

% ============================================================================
\onecolumn
\appendix
\section{Dataset Split Details}

\begin{table}[H]
\centering
\caption{K-RADAR sequence-level split. $^\dagger$Zero-shot weather (unseen during training).}
\label{tab:dataset}
\small
\begin{tabular}{llrrrlll}
\toprule
\textbf{Split} & \textbf{Seq} & \textbf{Frames} & \textbf{Obj} & \textbf{Obj/Fr} & \textbf{Weather} & \textbf{Road} & \textbf{Time} \\
\midrule
Train & 1   & 597  & 1{,}454 & 2.4 & Normal & Urban   & Night \\
      & 2   & 462  & 192     & 0.4 & Normal & Highway & Night \\
      & 3   & 597  & 1{,}365 & 2.3 & Normal & Highway & Night \\
      & 4   & 588  & 576     & 1.0 & Normal & Highway & Night \\
      & 6   & 594  & 234     & 0.4 & Normal & Urban   & Night \\
      & 8   & 567  & 1{,}440 & 2.5 & Normal & Univ.   & Night \\
      & 9   & 833  & 4{,}582 & 5.5 & Normal & Highway & Day   \\
      & 11  & 1{,}195 & 8{,}786 & 7.4 & Normal & Highway & Day   \\
      & 14  & 595  & 406     & 0.7 & Normal & Urban   & Day   \\
      & 21  & 597  & 5{,}112 & 8.6 & Rain   & Alleyway & Night \\
      & 28  & 597  & 597     & 1.0 & Sleet  & Mountain & Day   \\
      & 47  & 266  & 339     & 1.3 & Hvy.\ snow & Highway & Night \\
\cmidrule{2-8}
      & \multicolumn{1}{r}{\textit{12 seqs}} & \textit{7{,}491} & \textit{25{,}083} & \textit{3.3} & & & \\
\midrule
Val   & 5   & 597  & 3{,}386 & 5.7 & Normal & Urban   & Day   \\
      & 7   & 595  & 1{,}912 & 3.2 & Normal & Alleyway & Night \\
      & 23  & 598  & 1{,}735 & 2.9 & Rain   & Urban   & Night \\
\cmidrule{2-8}
      & \multicolumn{1}{r}{\textit{3 seqs}} & \textit{1{,}790} & \textit{7{,}033} & \textit{3.9} & & & \\
\midrule
Test  & 18  & 594  & 2{,}212 & 3.7 & Normal & Urban   & Day   \\
      & 38  & 597  & 558     & 0.9 & Fog$^\dagger$ & Mountain & Day   \\
      & 42  & 598  & 990     & 1.7 & Lt.\ snow$^\dagger$ & Urban & Day \\
      & 46  & 598  & 1{,}302 & 2.2 & Hvy.\ snow & Highway & Night \\
\cmidrule{2-8}
      & \multicolumn{1}{r}{\textit{4 seqs}} & \textit{2{,}387} & \textit{5{,}062} & \textit{2.1} & & & \\
\bottomrule
\end{tabular}
\end{table}

\section{Qualitative Results}

\begin{figure}[H]
\centering
\includegraphics[width=0.85\textwidth]{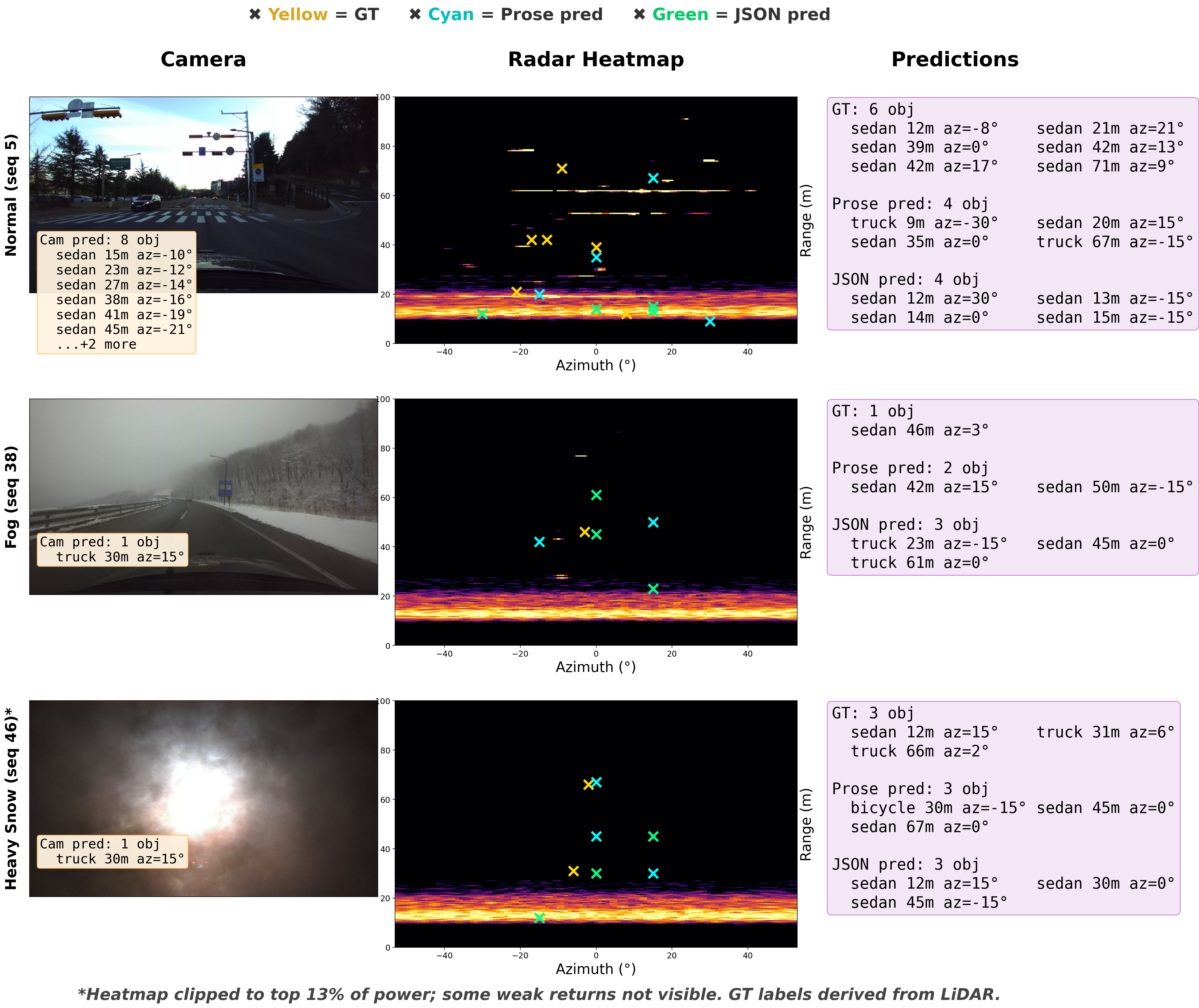}
\caption{Qualitative comparison across weather conditions.
Columns: camera image, radar range-azimuth heatmap, parsed caption summaries.
Rows: Normal (seq 5), Fog (seq 38), Heavy Snow (seq 46).
Yellow~$\times$\,=\,GT, cyan~$\times$\,=\,prose predictions,
green~$\times$\,=\,JSON predictions.
Radar RA maps remain interpretable in fog and heavy snow where camera
imagery is saturated or obscured.}
\label{fig:qualitative}
\end{figure}

\end{document}